\documentclass[runningheads]{llncs}
\usepackage{graphicx}
\usepackage[hidelinks]{hyperref}
\usepackage[utf8]{inputenc}
\usepackage{amsmath}
\usepackage{booktabs}
\usepackage[inline]{enumitem}
\usepackage{eurosym}

\begin{document}

\title{Contestable Black Boxes}
%
%

\author{Andrea Aler Tubella\inst{1}\orcidID{0000-0002-8423-8029} \and
Andreas Theodorou\inst{1}\orcidID{0000-0001-9499-1535} \and
Virginia Dignum\inst{1}\orcidID{0000-0001-7409-5813} \and
Loizos Michael\inst{2}}
\authorrunning{A. Aler Tubella et al.}
%
\institute{Umeå University, Sweden\\ 
\email{\{andrea.aler,andreas.theodorou,virginia.dignum\}@umu.se}\and
Open University of Cyprus \&
Research Center on Interactive Media,
Smart Systems, and Emerging Technologies, Cyprus\\
\email{loizos@ouc.ac.cy}}

\maketitle

\begin{abstract}  
The right to contest a decision with consequences on individuals or the society is a well-established democratic right. Despite this right also being explicitly included in GDPR in reference to automated decision-making, its study seems to have received much less attention in the AI literature compared, for example, to the right for explanation. This paper investigates the type of assurances that are needed in the contesting process when algorithmic black boxes are involved, opening new questions about the interplay of contestability and explainability. We argue that specialised complementary methodologies to evaluate automated decision-making in the case of a particular decision being contested need to be developed. Further, we propose a combination of well-established software engineering and rule-based approaches as a possible socio-technical solution to the issue of contestability, one of the new democratic challenges posed by the automation of decision making. 

\keywords{Right to contest  \and AI ethics \and Explainable AI }
\end{abstract}

\section{Introduction}

Searching for efficiency and cheaper solutions, governments and organisations are increasingly investing in automated solutions for a variety of decisions and activities, ranging from deciding on benefit claims to assessing the risk of recidivism in felons. 
With such life-changing determinations being treated automatically, the \textit{right to contest} a decision must be ensured for all automated decision applications.
Precisely outlining this necessity, Article 22 of the European Union's General Data Protection Regulations (GDPR) stipulates that whenever a decision which legally or significantly affects an individual relies solely on automated processing, then the right to contest the decision must be guaranteed. Similarly, although this right is not as explicitly phrased in the American law, we have already seen legal cases where the plaintiffs sued governmental organisations using algorithmic decision systems looking for accountability.

Complementing the current attention in the literature on \textit{fairness}, \textit{transparency}, \textit{explainability} and \textit{accountability} for automated decision-making systems, in this work we focus on \textit{the right to contest decisions}, an aspect that has received considerably less research focus.
This paper investigates the type of assurances that are needed in the contesting process when algorithmic black boxes are involved, opening new questions about the interplay of contestability and explainability. Further, we propose a combination of well-established software engineering and rule-based approaches as a possible socio-technical solution to the issue of contestability, one of the new democratic challenges posed by the automation of decision making.

\section{A Right in its Own Right}

The right to contest a decision that strongly impacts an individual takes many forms. It encompasses the right of appeal allowing to ask for a court’s decision to be changed and even the right to challenge the outcome of an election. Appealing is the democratic mechanism enacted to correct errors and allow for reparation.
In the case of automated decision-making, this right is equally acknowledged. For example, Article 22 of GDPR explicitly states the right to contest significant decisions relying solely on automated processing. 
Likewise, due process is a right recognised in the Anglo-American legal system, and it has been argued that individuals affected by decisions based on predictive algortihms should have similar rights to those in the legal system with respect to how their personal data is used in such adjudications, including the right to challenge them \cite{Crawford2013}. 

The ability to guarantee the right to contest is inextricably tied to the fundamental principles of \textit{responsible AI} \cite{Dignum2019ResponsibleWay}: transparency, explainability and accountability.
Transparency plays a crucial role in ensuring that stakeholders are aware that they are being subjected to automated decision making and that they have the right to challenge it \cite{Citron2014}. Further, to be able to understand a decision and assess if they believe that a contest needs to be raised, stakeholders need to be given a good explanation on how that decision was reached \cite{Wachter2017}. However, general provisions for explainability and transparency are not sufficient to guarantee the right of contest:
 whereas explanation and transparency methodologies focus on exposing the internal logic behind an algorithm or on describing how or why a specific decision was taken \cite{Miller2018}, they do not specifically reveal whether relevant rules and regulations have been adhered to or violated and why. To base this determination on an explanation requires a thorough examination of the explanation itself in light of the legal framework, which may be open to interpretation depending on its accuracy and level of detail. In contrast, the focus of a contesting procedure is expressly to ascertain \textit{post hoc} whether relevant rules and regulations were followed for a particular decision. Contesting therefore goes beyond the scope of explanation: it is not only the decision itself, but also the socio-legal context in which it was taken that need to be accounted for.
 
 To effectively address the right to contest a decision, policies should openly specify the regulations and adequacy determinations for specific applications \cite{Kroll2018}. Further, if a decision is challenged and mistakes are discovered a proper attribution of accountability and effective methods of compensation are needed. For this, AI governance is necessary to ensure that any \textit{moral responsibility} or \textit{legal accountability} is properly appropriated by the relevant stakeholders, together with the processes that support the redressing, mitigation, and evaluation of potential harm, alongside with the means to monitor and intervene on the system's operation. These should be accompanied with pre-established procedures for when a decision is contested, allowing to determine whether the relevant stipulations were followed in a way that is not open to interpretation. Thus, we argue
that specialised complementary methodologies to evaluate automated decision-making in the case of a decision being contested need to be developed.

For this reason, in the remainder of this paper we put forward a socio-technical approach to establishing a contesting procedure for automated decisions, combining well-established software engineering practices and rule-based approaches. We propose that any contestable automated system should be accompanied by a formal specification that describes in an unambiguous language the constraints that each constituent agent-module of the system needs to fulfil. This formal specification effectively corresponds to the organisation’s \textit{legal-compliance contract},  and includes all necessary legal and consumer-protection requirements. When a system's decision is contested in a certain context, this is taken as a request to verify that the system indeed operated in line with its accompanying formal specification in that particular context. Verification is achieved by examining the system's (and each agent-module's intermediate) behaviour while monitoring for the violation or fulfilment of the constituent provisions of the specifications.

Since the contesting procedure is an examination of a decision already taken,  software development practices that ensure \textit{traceability} become fundamental. Furthermore, to ensure that all the relevant factors for the review of the decision are being preserved, this should be done in conjunction with standardised \textit{elicitation} and \textit{interpretation} processes to identify the relevant policy that the system is mandated to adhere to and to translate it into specific  constraints on the system. These requirements are part of designing intelligent systems responsibly\cite{Dignum2019ResponsibleWay}, and align with the push for relating high-level governance, including legal and ethical considerations, with concrete system functionalities \cite{VandePoel2013}.

\section{Compliance Contract}\label{management}

We propose that any contestable automated system should be accompanied by a formal specification that describes in an unambiguous language the constraints that each constituent agent-module of the system needs to fulfil. Our proposed approach includes the following steps: \begin{enumerate*}
    \item \textit{norm management},
    \item \textit{norm formalisation}, and
    \item \textit{negotiation and validation}.
\end{enumerate*}
Each of these stages is further broken down into explicit steps, described in this section. The end-product of this process is the organisation's compliance contract, which can then be used for monitoring specific decisions in case of contest. 

To illustrate our method, we will use the real-life example of Lufthansa's automated pricing algorithm, which increased prices up to 30\% immediately following the bankruptcy of competitor Air Berlin \cite{Bundeskartellamt2018NoPricing}. Following consumer complaints of abusive monopoly, the German consumer-protection regulator, Bundeskartellamt, conducted an investigation. Lufthansa’s initial response was that the algorithm acted autonomously, but Bundeskartellamt made it clear that even though Lufthansa was cleared of wrongdoing, the fact that price increases were the result of an automated algorithm had no bearing on their decision. 

\paragraph{\textbf{Norm management.}} Building concrete specifications for a compliance contract starts, similarly to a `traditional' software life cycle, with the \textit{norm management phase}. Inspired by the \textit{IEEE Recommended Practice for Software Requirements Specification} \cite{IEEE830}, we propose a two-step process: \textit{elicitation} followed by \textit{interpretation}. Each of these phases necessarily involves the participation of both the software development team and the legal department of the organisation, whose different areas of expertise will be fundamental in obtaining norms that are both implementable and legally sound. 

The \textbf{elicitation stage} takes place by consulting governance, i.e. standards and legislation relevant to the system. The purpose of this stage is not to set the norms at once, but rather to identify and list the relevant policy that the system is mandated to adhere to. The produced list of rules and guidelines provide the high-level policy that not only the software deliverable itself, but also its development, deployment, and usage processes need to follow. In the case of an airline pricing system, relevant laws include the anti-monopoly and consumer-protection regulations. Moreover, laws such as non-discrimination should also be followed by the system. For example, in the case of Lufthansa's pricing algorithm, German anti-monopoly laws clearly apply. 

The standard practice for setting airline ticket prices is to have similar seats divided in tiers, where each tier is in a more expensive preset price range than the previous one. The first seats sold belong to the first tier, and only when all of them are sold the more expensive seats of the second tier are made available to buy. Thus, price steadily increases as the plane fills up. In the case of a competitors' bankruptcy, airlines are not allowed to capitalise on it by imposing extreme price increases or selling only the most expensive tiers. Lufthansa was cleared from wrongdoing by demonstrating two points: 
\begin{enumerate*} 
\item[(1)] that only the comparatively more expensive booking tiers were available as the cheaper booking classes were imminently booked, and
\item[(2)] that the price range for each tier was comparable to previous years' prices and not illegally increased.
\end{enumerate*} 
These requirements set out by Bundeskartellamt would be clearly identified at this stage, and set down as the basis for the compliance contract.

In order to interpret these rules into concrete checkable norms, abstract concepts such as \textit{``comparable to previous years' prices"} must be turned into concrete computable requirements. Thus, the \textbf{interpretation stage} entails a translation of high-level governance and legal requirements into concrete norms specifying the constraints that the system needs to fulfil, taking into account its purpose and the context of its deployment. The resulting norms should be comprehensive enough so that fulfilling them can prove that the system is adhering to the ethical and legal policy of its developers' organisation. The shift from abstract to concrete necessarily involves careful consideration of the context identified in the previous phase. In this sense, the implementation of each requirement will vary from context to context the same way it can vary from system to system. In our simple example, an acceptable set of concrete norms that the company could set that would satisfy the consumer protection agency's concerns over the anti-monopoly regulations would be given by:
 \begin{enumerate*} 
\item[(1)]``cheaper tiers must be fully booked before more expensive tiers are made available'' and 
\item[(2)]``the pricing range of a tier does not differ by more than 30\% from the average price of the same tier on the same route on the same day in the previous 5 years''.
\end{enumerate*}

\paragraph{\textbf{Formalisation}}\label{formalisation}
This phase entails the formal specification of the constraints for each of the component agents. This step requires a further concretisation of the norms: they need to be formulated in a way that makes them operational and allows for the detection of violations. This requires linking the concepts contained in the rules with a rule-based language that will determine the normative framework \cite{Aldewereld2007}. The formal normative system obtained in this stage will constitute the organisation's proposed \textit{compliance contract}.

An appropriate language for this purpose needs sufficient expressiveness to model the relevant legal requirements and policies, while allowing for compliance with these requirements to be monitored and tracked. 
Many existing approaches to norm monitoring in the MAS literature take an \textit{enforcement} point of view, in which a monitoring system is an \textit{observation} mechanism, that can log norm violations provided it can access the relevant information \cite{King2017,Modgil2009}. This mechanism is particularly developed for scenarios where each participant and component of the system has a well-defined purpose (such as buyer or seller) with clear actions available for each role (such as buy, sell, negotiate, concede) and has been successfully applied in contexts like marketplaces \cite{Michael2010}. This type of encoding is therefore a proposed approach in this type of well-bounded scenario.
 
 In heavily regulated scenarios, powerful expressive languages to capture the semantics of the regulations and their interactions are needed. To monitor adherence, these must be combined with a useful representation of the computation and run-time events of the decision being observed. A similar challenge is found in the case of \textit{automated compliance checking} of business processes. Languages for automated compliance checking provide both a formalism to model and reason with regulations, and formal representations of processes \cite{Governatori2016,DeVos2019}. Our proposed contesting procedure, likewise, requires monitoring adherence to regulations, but for determined and completed instances of a computation rather than for verifying adherence of processes. For this reason, we propose that automated compliance checking methods and languages could be usefully adapted to this framework.



\paragraph{\textbf{Negotiation and validation}}\label{negotiation}
This stage facilitates an open discussion of the proposed compliance contract with representatives of all groups of internal stakeholders; from product managers to software developers to the quality assurance engineers to legal experts. Where possible, external stakeholders---such as users and regulators---should also be consulted.
These discussions aim to validate the norms to ensure their accurate interpretation of relevant legislation and their acceptance. The approval of the relevant regulators is particularly desirable, as their acceptance of the norms as a compliance contract entails that showing that the norms where adhered to is enough to dismiss a complaint under the grounds that the decision is fully legal. If possible, a negotiation and validation phase would occur after each phase of the process of obtaining the compliance contract, to maintain maximum transparency.

\section{Examining a Contested Decision Under Monitoring}\label{contest}


To adequately examine the original decision-taking under the compliance contract, both the inputs that the target process received and the `state of the world' that held when the original decision was taken must be recorded. In the same vein, depending on the constraints imposed by the norms, the occurrence of certain events will need to be tracked and recorded as well. In the case of our example, knowing when a tier has been sold out, knowing which prices have been set for each tier through time and when seats from a certain tier were put on sale is indispensable to ascertain whether the criteria set by Bundeskartellamt were followed. Awareness of this trace of events is fundamental whether it is a human or an artificial agent that makes the determination of whether rules were followed. 
Furthermore, if the decision-taking algorithm is adaptive (for example learning from new data and adapting its behaviour) then it is the version that held at the time of the original decision that should be evaluated. Both of these challenges can be addressed with version control and thorough record keeping. Like formal specifications, version control is part of software engineering: even for machine learning approaches, advanced forms of version control including record keeping of data is recommended and increasingly used \cite{Bryson2019}.
This practice ensures \textit{traceability} and, therefore, reproducibility and auditability --as in the context of a contested decision.

The compliance contract could of course be used to check every decision for compliance with the specifications, or even to forcibly enforce adherence to norms.  For example, the system could be endowed with norm-reasoning mechanisms forcing it to act upon the specified norms \cite{Jensen2014}. This could however considerably slow down the computation, and be expensive in terms of resources. Indeed, at each decision and action, the system would need to check whether a norm applies, and then how to act upon it. To assess the former, it may even need to access extraneous information about the state of the world. If an organisation is willing to pay this cost, this regimentation approach could be deployed for every decision, or perhaps for every \textit{critical} decision as a safeguard. However, it may be preferable cost-wise to use monitoring specifically in the case of contests, and bear the cost of sanctions instead when norms have been violated.

\section{Discussion and Future Work}

Whenever a decision process takes place that has individual or social consequences, the right to contest the said decision is a well-established democratic right. 
In this paper we focused on \textit{contestability} of decisions through the application of well-established software engineering and rule-based policy modeling techniques. 

An important requirement for our proposed approach is that the norms identified in the elicitation and interpretation stages should be captured accurately in a computational language that can be used for specification and automated monitoring. If the normative framework is very complex, such as cases where great knowledge about the state of the world is required or when reasoning about causes and consequences is necessary, this can become a challenge. Research on how to completely capture regulatory frameworks is ongoing, involving the fields of policy modeling, normative reasoning and knowledge representation amongst others. For this reason, we expect this approach to work best in cases where the regulation is very clear and focused on the behaviour of the system itself, with limited dependence on the outside world. The range of application of our proposed approach will keep increasing,  as more approaches are developed for increasingly complex normative frameworks.

An additional cornerstone of our proposed approach is the requirement of exhaustive record-keeping, to make decisions examinable. Although such good software development practices should be standard, they could prove technically challenging for some applications, or could interfere with other requirements such as data protection and privacy. Our proposed approach is versatile enough to still be applicable in such cases: by re-computing the decision under the monitoring agent, rather than operating with a record of events, certain  norm violations could still be identified.

We believe our proposal opens new interesting research questions for further examination. First and foremost, it promotes a new avenue of research for rule-based representations of complex norms. While we have only begun to consider monitoring approaches, there is a need to conduct real-world implementations to identify real-world needs. Furthermore, looking beyond the implementation details, we call upon an exploration on how contestability and explainability differ in terms of costs, system requirements and trust calibration for naive and expert users, and advocate the development of concrete methodologies oriented specifically to contestability.

\section*{Acknowledgements}

A. Theodorou was supported by the European Union’s Horizon 2020 research and innovation programme under grant agreement no. 825619 (AI4EU project). 

A. Aler Tubella and V. Dignum were supported by the Wallenberg AI, Autonomous Systems and Software Program (WASP) funded by the Knut and Alice Wallenberg Foundation.

L. Michael was supported by funding from the EU’s Horizon 2020 Research and Innovation Programme under grant agreements no. 739578 and no. 823783, and from the Government of the Republic of Cyprus through the Directorate General for European Programmes, Coordination, and Development.


\bibliographystyle{splncs04}
\bibliography{contest}  

\end{document}